\documentclass{article}

\usepackage{microtype}
\usepackage{graphicx}
\usepackage{subfigure}
\usepackage{booktabs} 

\usepackage{hyperref}

\usepackage[accepted]{icml2025}
\usepackage{enumitem} 
\usepackage{amsmath}
\usepackage{amssymb}
\usepackage{mathtools}
\usepackage{amsthm}

\usepackage[capitalize,noabbrev]{cleveref}

\theoremstyle{plain}

\theoremstyle{definition}

\theoremstyle{remark}

\usepackage[textsize=tiny]{todonotes}

\icmltitlerunning{SatFlow: Generative model based framework for producing High Resolution Gap Free Remote Sensing Imagery.}

\begin{document}

\twocolumn[
\icmltitle{SatFlow: Generative Model based Framework for Producing High Resolution Gap Free Remote Sensing Imagery.}




\begin{icmlauthorlist}
\icmlauthor{Bharath Irigireddy}{usda}
\icmlauthor{Varaprasad Bandaru}{usda}
\end{icmlauthorlist}

\icmlaffiliation{usda}{Agricultural Research Service, United States Department of Agriculture, Maricopa, AZ, USA}

\icmlcorrespondingauthor{Varaprasad Bandaru}{prasad.bandaru@usda.gov}
\icmlcorrespondingauthor{Bharath Irigireddy}{bharath.irigireddy@usda.gov}

\icmlkeywords{Remote Sensing, Machine Learning, ICML}

\vskip 0.3in
]



\printAffiliationsAndNotice{}  

\begin{abstract}

Frequent, high-resolution remote sensing imagery is crucial for agricultural and environmental monitoring. Satellites from the Landsat collection offer detailed imagery at 30m resolution but with lower temporal frequency, whereas missions like MODIS and VIIRS provide daily coverage at coarser resolutions. Clouds and cloud shadows contaminate about 55\% of the optical remote sensing observations, posing additional challenges. To address these challenges, we present SatFlow, a generative model based framework that fuses low-resolution MODIS imagery and Landsat observations to produce frequent, high-resolution, gap-free surface reflectance imagery. Our model, trained via Conditional Flow Matching, demonstrates better performance in generating imagery with preserved structural and spectral integrity. 
Cloud imputation is treated as an image inpainting task, where the model reconstructs cloud-contaminated pixels and fills gaps caused by scan lines during inference by leveraging the learned generative processes. Experimental results demonstrate the capability of our approach in reliably imputing cloud-covered regions. This capability is crucial for downstream applications such as crop phenology tracking, environmental change detection etc.,
\end{abstract}

\section{Introduction}

High spatial and temporal resolution remote sensing imagery enables a wide range of agricultural and environmental monitoring applications, including phenology mapping, yield forecasting, and meteorological disaster prediction \cite{BOLTON2020111685, Gillespie2007Assessment, Huber2024Leveraging}. Optical remote sensing imagery provides rich spectral information with strong interpretability. The Landsat program, operational since 1972, provides decades of Earth observation data at 30 m spatial resolution, enabling detailed land surface monitoring over an extended period. However, infrequent revisit intervals (10-16 days) and data gaps caused by cloud cover during imaging and the Scan Line Corrector failure in Landsat 7 pose significant challenges to consistent monitoring \cite{zhu2012fusion}. Cloud contamination is of particular concern, affecting up to 55\% of optical remote sensing observations over land globally \cite{King2013Spatial}, leading to substantial loss of clear-sky scenes and limiting subsequent image analysis.
These issues are especially acute in agricultural regions, where landscapes are highly dynamic during growing season and high temporal frequency is critical for capturing rapid changes in vegetation growth and phenological transitions. On the other hand, the Moderate Resolution Imaging Spectroradiometer (MODIS) instruments aboard NASA's Terra (launched in 1999) and Aqua (launched in 2002) satellites provide near-daily global coverage at resolutions ranging from 250m to 1km \cite{Xiong2009NASA}. While this temporal fidelity is ideal for tracking short-term changes, the coarse resolution is insufficient for capturing field-level agricultural details or fine-grained ecosystem processes. Nevertheless, the MODIS record, spanning over two decades, forms an invaluable resource for environmental applications, including forest cover change monitoring, urban expansion mapping, and wildfire impact assessment \cite{Liu2024GlobalTree, SCHNEIDER20101733}. Integrating MODIS's rich temporal information with Landsat's fine spatial detail offers an opportunity to generate a spatiotemporally enhanced long-term dataset that can inform a broad range of land surface and environmental monitoring and modelling applications.

\begin{figure}[ht]
\vskip 0.2in
\begin{center}
\centerline{\includegraphics[width=\columnwidth]{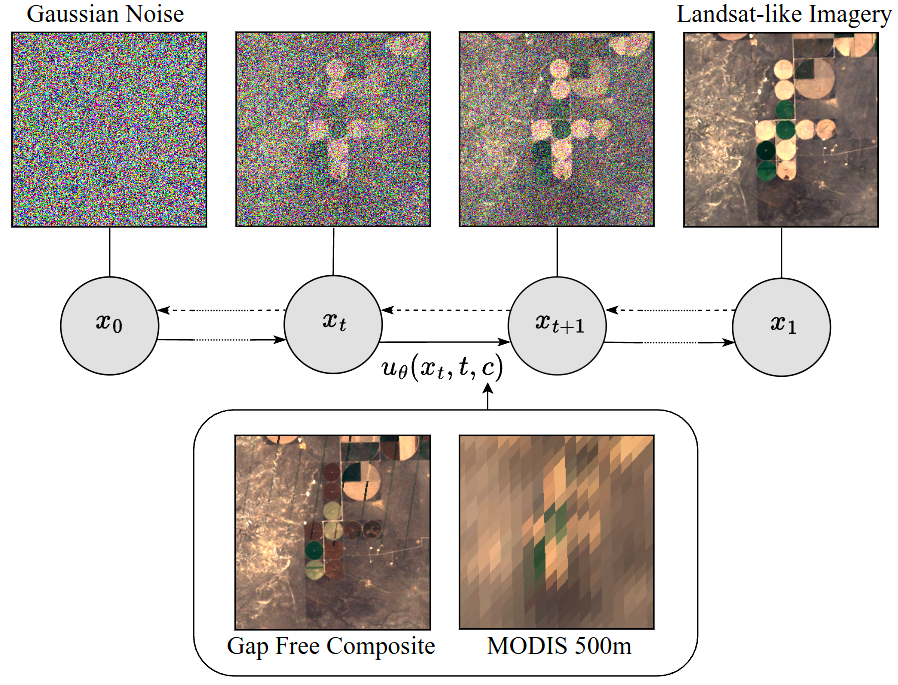}}
\caption{The framework integrates MODIS and Landsat observations through conditional flow matching to downscale MODIS imagery (500m) to Landsat resolution (30m).}
\label{fig:framework}
\end{center}
\vskip -0.2in
\end{figure}


Several approaches have been investigated to achieve such spatiotemporal integration. Established fusion methods—such as the 
Spatial and Temporal Adaptive Reflectance Fusion Model (STARFM) \cite{gao2006starfm}, the SpatioTemporal Adaptive fusion 
of High-resolution satellite sensor Imagery (STAIR) \cite{zhu2010fusion}, and the Highly Integrated STARFM (HISTARFM) 
\cite{zhu2016histarfm}—blend temporally frequent but coarse imagery with sparse but fine-resolution 
observations. While these methods have demonstrated improvements, they often encounter challenges in heterogeneous landscapes and during periods of rapid land-cover change. STARFM and its variants are limited by the need to manually select one or more suitable pairs of coarse and high-resolution images for each fusion task, which poses challenges for automation at scale.

Advances in machine learning and deep generative models, including Generative Adversarial Networks (GANs) \cite{goodfellow2014generative} and diffusion-based approaches \cite{ho2020denoising} have shown promise in image synthesis and super-resolution tasks  \cite{Wang2019, Lim2017EnhancedDeepSR, Diffusion22}. While GANs can yield highly realistic imagery, they may suffer from training instability and spectral inconsistencies \cite{dhariwal2021diffusion}.
Few works have applied generative models to remote sensing domain \cite{Xiao2024EDiffSR, khanna2024diffusionsat} and these typically require a large number of inference steps, as noted by Zou et.al.\yrcite{DiffCR2024}. While Zou et.al. proposed an efficient diffusion approach for cloud imputation, it is limited to static landscapes and it can not be adapted to dynamic agricultural environments. Our novel framework integrates MODIS observations for contextual information while gap-filling high-resolution imagery. Beyond GANs and diffusion, our work utilizes Conditional flow matching \cite{Lipman2023Flow,Tong2024Improving}, a growing class of generative models that allow for exact likelihood estimation and often exhibit more stable training. The key contributions of our work are: (1) We present a novel approach for downscaling coarser-resolution MODIS imagery using a generative model to synthesize Landsat-like imagery. (2) We propose a gap-filling strategy that leverages the learned generative process to fill missing pixels in Landsat observations caused by cloud cover and scan lines. (3) We integrate the model into a pipeline to generate high-resolution, gap-free Landsat-like imagery at regular intervals.

\section{Methodology}

\subsection{Flow Matching Formulation}

The primary objective is to generate gap-free surface reflectance images given the conditioning factors, 
which include corresponding low-resolution MODIS imagery and a gap-free composite of previously 
acquired Landsat images. Our framework builds on conditional flow matching \cite{Lipman2023Flow,Tong2024Improving}, 
which generalizes continuous normalizing flows \cite{Grathwohl2019FFJORD,NODE2019} by directly regressing the vector fields for transforming between noise and data distributions.  
The goal of flow matching, similar to diffusion models \cite{ho2020denoising, Diffusion22}, 
is to generate samples that lie in the data distribution through an
iterative process. We refer to the starting random gaussian noise distribution as $x(0)$ and the 
gap-free Landsat data distribution as $x(1)$, where the generative modeling task is to transform 
the initial noisy input $x_0$ to the target distribution $x_1$, through a learned process that is
guided by the conditioning factors $c$ (illustrated in \cref{fig:framework})    . 

\subsection{Training}
\label{subsec:model_training}
To learn a model that can transform $x(0)$ to $x(1)$, we model
a time-varying vector field $u(t): [0, 1] \times \mathbb{R}$, defined by the following ordinary differential
equation: $u(t) = {dx(t)}/{dt}$. and a probability path $p(t) : [0, 1] \times \mathbb{R}$.
 Intuitively, this vector field defines the direction and magnitude by 
which to move a sample in $x_0$ so that it arrives at its corresponding location in $x_1$ by following 
the probability path $p$ over time. We aim to approximate the true vector field $u$ using a neural 
network $u_{\theta}(x_t, t, c)$, parameterized by weights $\theta$. The flow matching objective is to
minimize the difference between the predicted vector field $u_{\theta}(x_t, t, c)$ and the true vector
field $u_t$, as expressed in \cref{eq:flow_matching}:

\begin{equation}
\label{eq:flow_matching}
\min_{\theta} \mathbb{E}_{t, x_t \sim p(x_t|x_0, t)}\left[\|u_{\theta}(x_t, t, c) - u_t\|^{2}\right].
\end{equation}

However, this objective is intractable as there is no closed
form representation for the true vector field $u(t)$. Instead, similar to approaches that leverage simple
linearized paths for training \cite{Liu2023FlowStraight,Pooladian2023Multisample},
we model the vector field $u_t$ and the probability path $p: [0, 1] \times R$ between $x_0$ and $x_1$ 
with standard deviation $\sigma$ as shown in \cref{eq:vector_field,eq:prob_path}. 

\begin{equation}
\label{eq:vector_field}
 u(t) = x_1 - x_0
\end{equation}
\begin{equation}
\label{eq:prob_path}
 x_t \sim \mathcal{N}( (1 - t)\cdot x_0 + t\cdot x_1, \sigma^2)
\end{equation}

\Cref{eq:prob_path} defines the probability path as a Gaussian distribution
centered at a linear interpolation between $x_0$ and $x_1$ at time $t$.
\Cref{eq:vector_field} defines the target vector field
simply as the difference vector pointing from the starting
point $x_0$ to the end point $x_1$. The training procedure to approximate this vector field is outlined in Algorithm \ref{alg:training}.

\begin{algorithm}[H]
\caption{Conditional Flow Matching Training}
\label{alg:training}
\begin{algorithmic}[1]
\REQUIRE initial parameters $\theta$, learning rate $\alpha$ 
\REPEAT
    \STATE Sample a batch of final states $x_1$, corresponding conditions $c$,
        initial states $x_0 \sim \mathcal{N}(0, I)$ and $t \sim [0, 1]$.
    \STATE Compute the true vector fields: $u_{t} = x_1 - x_0$ 
    \STATE Sample $x_t \sim \mathcal{N}\bigl((1 - t)\cdot x_0 + t \cdot x_1, \text{ }\sigma^2\bigr)$
    \STATE Compute the loss: \\
     \quad \quad  $L_{CFM}(\theta) = \frac{1}{2} \| u_{\theta}(x_t, t, c) - u_{t}\|^{2}$
    \STATE Update parameters:
    $\theta \leftarrow \theta - \alpha \nabla_{\theta}L_{CFM}(\theta)$
\UNTIL{converged}
\end{algorithmic}
\end{algorithm}

In algorithm \ref{alg:training}, $x_1$ represents ground-truth Landsat imagery, while the conditioning factors $c$ consist of two components: (1) MODIS observations acquired on the same date as $x_1$, providing coarse-resolution spectral information, and (2) a gap-free composite constructed from previously captured Landsat images of the same scene, providing high-resolution spatial context. Ideally, the model has to learn to synthesize Landsat-like high-resolution imagery by jointly leveraging the spatial structure from the composite and the spectral characteristics from MODIS data. To achieve this, we employ two key strategies during the training process: (1) MODIS inputs are randomly masked with a probability of $50$ \%, and (2) the gap-free composite is randomly selected from multiple available composites of the same scene (illustrated in Figure \ref{fig:Training}). This augmentation approach encourages the model to disentangle and effectively utilize both information sources - learning to preserve spatial details from the composite while imparting the spectral information from MODIS observations when available. During inference, if MODIS observations are unavailable, the model generates plausible, unconditional spectral signatures, while still conforming to the spatial characteristics of the scene dictated by the Landsat composite.

\subsection{Inference}

To generate a sample from the target distribution $x_1$, given conditioning factors $c$, we integrate the learned vector field $u_{\theta}(x_t, t, c)$ over time. Specifically, starting from an initial sample $x_0 \sim \mathcal{N}(0, I)$, we follow the trajectory defined by \cref{eq:trajectory}:
\begin{equation}
\label{eq:trajectory}
x_1 = x_0 + \int_{0}^1 u_{\theta}(x_t, t, c) dt 
\end{equation}
In practice, we approximate this integral using a discrete-time numerical scheme \cite{NODE2019, Lipman2023Flow}. Forward Euler approach is employed 
for simplicity and computational efficiency as shown in \cref{alg:inference}.

\begin{algorithm}[H]
\caption{Conditional Flow Matching Inference}
\label{alg:inference}
\begin{algorithmic}[1]
\REQUIRE conditions $c$, time step $dt$,  initial $x_0 \sim \mathcal{N}(0, I)$, 
\FOR{$t = 0$ to $1$ \textbf{step} $dt$}
    \STATE $x_{t + dt} = x_{t} + u_{\theta}(x_t, t, c) \cdot dt$
\ENDFOR
\OUTPUT $x_1$
\end{algorithmic}
\end{algorithm}

In \cref{alg:inference}, starting from a random noise distribution, the model iteratively updates $x_t$ using the vector field $u_{\theta}(x_t, t, c)$ to produce the final high-resolution Landsat-like imagery $x_1$. Clouds and scanlines imputation approach in our framework is inspired by the image inpainting methodology investigated by Lugmayr et al. \yrcite{lugmayr2022repaint} in the context of diffusion models.
Given clouds or scan line contaminated images and their corresponding quality assessment mask $m$ (where $m_i=1$ indicates cloudy/missing pixels and $m_i=0$ indicates clear pixels), we introduce a composite update strategy that
relies on the learned vector field $u_{\theta}(x_t, t, c)$ to reconstruct the unknown pixels and for the known pixels, uses a direct interpolation with the observed values as shown in \cref{alg:imputation}.

\begin{algorithm}[H]
\caption{Cloud Imputation and Scan Lines Filling}
\label{alg:imputation}
\begin{algorithmic}[1]
\REQUIRE cloudy images $x_1^*$, cloud mask $m$, conditions $c$, time step $dt$,  initial states $x_0 \sim \mathcal{N}(0, I)$, 
\STATE Compute: $u = x_1^* - x_0$
\FOR{$t = 0$ to $1$ \textbf{step} $dt$}
    \STATE $x_{t + dt} = x_{t} + ( u_{\theta}(x_t, t, c)\cdot m  + u\cdot(1 - m)) \cdot dt$
\ENDFOR
\OUTPUT $x_1$
\end{algorithmic}
\end{algorithm}

This composite strategy ensures physical consistency by respecting the known data where available while leveraging the learned generative processes to fill in missing regions.
\Cref{alg:inference,alg:imputation} demonstrate the versatility of the model in both generating high-resolution imagery and performing gap filling of the acquired imagery. 

\subsection{Model Architecture}

The model architecture employs a U-Net \cite{Ronneberger2015UNetCN} design augmented with ResNet-style blocks \cite{He2016DeepRL} and self-attention layers \cite{Vaswani2017AttentionIA, Bello2019attentionaugmented}, as illustrated in Figure 2. Conditioning information comprising MODIS observations and a gap-free Landsat composite - is concatenated along the channel dimension with the current state $x_t$. The current time step $t$ and ancillary metadata which includes day of year (DOY), sensor type (TM/OLI) and MODIS availability flag are encoded via learned embeddings and injected into the network at multiple resolutions. The network processes the inputs through a series of downsampling and upsampling stages linked by skip connections. Residual connections help stabilize training, and self-attention mechanisms capture both local and global dependencies. The network outputs the vector field $u_{\theta}(x_t, t, c)$ with six channels, corresponding to the multi-spectral dimensions of the Landsat data.

\begin{figure*}[!h]
    \vskip 0.2in
    \begin{center}
    \centerline{\includegraphics[width=\textwidth]{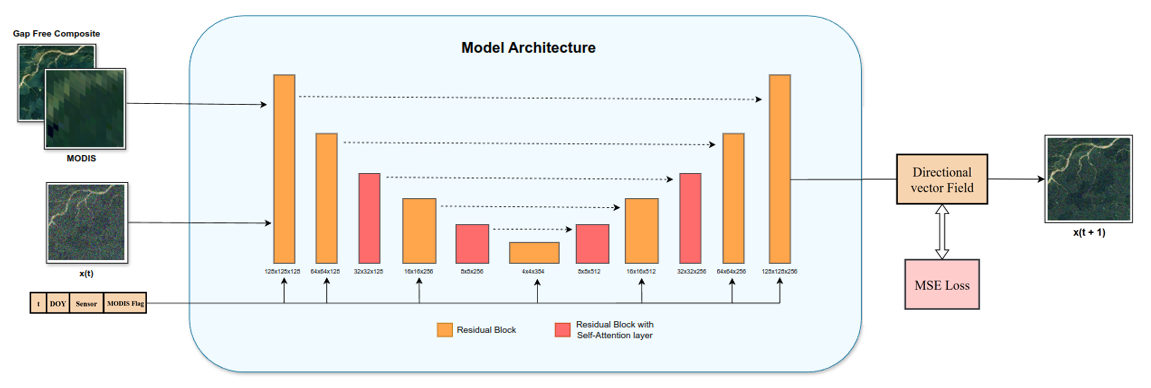}}
    \caption{The Conditioning input are concatenated along the channel dimension with the current state $x_t$. The current time step $t$ and metadata are encoded via learned embedding and integrated into the network at multiple resolutions. The network predicts the vector field $u_{\theta}(x_t, t, c)$ and MSE loss is computed between the predicted and target vector fields.}
    \label{fig:framework}
    \end{center}
    \vskip -0.2in
\end{figure*}

\subsection{Overall Framework}
We integrate the trained generative model into a pipeline to produce gap-free high resolution imagery at regular intervals. The framework processes two complementary data streams: daily MODIS imagery and Landsat observations (Landsat 5-9) with varying revisit times. The pipeline comprises of three key components: (1) Pre-processing of MODIS imagery: A temporal interpolation module that fills cloud-contaminated pixels in the MODIS time series using clear observations from adjacent days; (2) A gap-filling module that fills the clouds and scan-lines in the acquired Landsat scenes utilizing the trained model (\cref{alg:imputation}) (3) Finally, Landsat-like imagery are synthesized by the model at regular intervals by fusing the processed MODIS observations and gap-filled Landsat scenes. Since MODIS sensors (Aqua and Terra) acquire global imagery on a near-daily basis (as opposed to Landsat’s 2–6 observations per month), temporal interpolation allows short gaps to be reconstructed with minimal discrepancy. Gap-filling module leverages spatial context from the Landsat composite and spectral information from temporally rich MODIS observations. The hierarchical design of the framework enables robust spatio-temporal fusion. A similar approach can be adapted to integrate other remote sensing data sources (e.g., VIIRS, Sentinel-2, SAR) for broader applicability.

\section{Experiments}
\subsection{Dataset}

The dataset for training the model was derived from Landsat and MODIS satellite imagery, 
spanning the period from years 2000 to 2024 across the Contiguous United States (CONUS). 
The years 2012 and 2015 were chosen to be excluded from the training set for validation, as these years represent contrasting dry and wet conditions respectively. 
The study utilized Level 2 processed surface reflectance data from Landsat 5, 7, 8, and 9 missions \cite{CRAWFORD2023100103} and the 
MODIS Bidirectional Reflectance Distribution Function (BRDF)-corrected MCD43A4 product \cite{schaaf2002first, lucht2000algorithm}. 
The MCD43A4 product integrates data from the Moderate Resolution Imaging Spectroradiometer (MODIS) sensors aboard the Aqua (launched in 2002) and Terra (launched in 1999) satellites, which observe Earth's surface at different times during the day \cite{Link2017Status}, providing daily global coverage at $500m$ resolution. Using stratified sampling, $\mathbf{20,000}$ locations were sampled across the contiguous United States based on the Cropland Data Layer (CDL) of year 2020 provided by USDA-NASS, covering diverse land cover and crop types.
For each sampled location, imagery was obtained from four different dates where the cloud cover was less than ${10\%}$ amounting to $\mathbf{80,000}$ data points for training. The dataset included: (1) Landsat surface reflectance imagery of size $256 \times 256$ pixels at $30m$ resolution, containing the six spectral bands (Red, Blue, Green, near-infrared (NIR), and two shortwave infrared 
bands (SWIR1 and SWIR2)); (2) corresponding MODIS imagery, 
resampled and aligned to match Landsat's spatial resolution; and (3) gap-free composites generated by stacking temporally preceding Landsat scenes. These composites 
were created by applying quality assessment masks to eliminate clouds, scan lines, and cloud shadows, followed 
by mosaicking the remaining clear pixels to ensure continuous spatial coverage. The data processing 
and collection workflow was implemented on Google Earth Engine. Prior to training, the reflectance values are normalized to lie within $[-1,1]$ range using the scaling coefficients computed over the training set. 

\subsection{Setup}

The model was trained to minimize the Mean Squared Error (MSE) loss between the predicted and target vector fields, following the procedure outlined in section~\ref{subsec:model_training}. We adopted the AdamW optimizer \cite{loshchilov2019decoupled} with a base learning rate of $1e\text{-}4$. A cosine learning rate schedule \cite{loshchilov2017sgdr} with $6,000$ warmup steps was employed to improve convergence and mitigate potential instabilities during the early stages of training. Each training spans $120$ epochs and was conducted on two NVIDIA RTX A6000 GPUs, each processing a batch of $16$ images. We further applied gradient accumulation over $4$ steps, effectively increasing the batch size without exceeding GPU memory limits. All training runs employed a standard deviation of $\sigma = 0.001$ in Algorithm~\ref{alg:training} to define the probability path. Influence of alternative choices for standard deviation ($\sigma$) were not investigated in our work.

We validated our method on a dataset comprising 2,500 held-out scenes from 2012 and 2015. First, we evaluated downscaling quality (from 500m MODIS to 30m Landsat resolution), comparing the model's predicted Landsat-like outputs with the actual high-resolution Landsat images. we also trained the same model architecture with conditional diffusion methodology as outlined by Zou et al. \yrcite{DiffCR2024} to compare the performance with conditional flow matching. A sigmoid noise schedule rescaled to a zero terminal signal-to-noise ratio (SNR) is implemented for the diffusion model, as it demonstrated superior performance. we assess gap-filling performance by synthetically masking clean Landsat imagery with varying cloud coverage levels (10\%--75\%) using randomly generated cloud masks \cite{Czerkawski2023}. These artificial gaps are filled by the model as outlined in Algorithm~\ref{alg:imputation}, enabling direct comparisons against the known ground truth reflectance values. To our knowledge, no publicly available benchmarks are suitable for evaluating our data-fusion and cloud imputation approaches.

\subsection{Evaluation Metrics}
To assess the quality of the generated surface reflectance images, we employ the following metrics: 

\textbf{1. Spectral Information Divergence (SID):}  
Spectral Information Divergence \cite{Chang2000SID} is an information-theoretic metric introduced to measure discrepancies between two spectral signatures. 
In our evaluation, we compute SID across all six spectral 
bands (Red, Green, Blue, NIR, SWIR1, SWIR2) between the generated and original Landsat imagery. Lower SID values indicate that the reconstructed 
spectrum closely matches the reference. The SID between two spectral signatures $p$ and $q$ is given by:

\begin{equation}
SID(p,q) = \sum_{i=1}^{N} p_i \log\left(\frac{p_i}{q_i}\right) + \sum_{i=1}^{N} q_i \log\left(\frac{q_i}{p_i}\right)
\end{equation}

where $p_i$ and $q_i$ represent the normalized reflectance values for band $i$ in the generated and reference images respectively.

\textbf{2. Structural Similarity Index Measure (SSIM):}  
Structural Similarity Index Measure \cite{Wang2004SSIM} is computed over local $11\times11$ pixel windows. For each window pair $x$ and $y$:

\begin{equation}
SSIM(x,y) = \frac{(2\mu_x\mu_y + c_1)(2\sigma_{xy} + c_2)}{(\mu_x^2 + \mu_y^2 + c_1)(\sigma_x^2 + \sigma_y^2 + c_2)}
\end{equation}

where $\mu_x$, $\mu_y$ are the mean intensities of windows $x$ and $y$ respectively, $\sigma_x^2$, $\sigma_y^2$ are their 
variances, and $\sigma_{xy}$ is the covariance between the windows. 
The final SSIM score is obtained by averaging across all windows and RGB bands, 
higher values indicating greater structural similarity.

\textbf{3. Peak Signal-to-Noise Ratio (PSNR):}
Peak Signal-to-Noise Ratio is useful for evaluating the pixel-wise accuracy, with a typical range of 20 to 40 dB for acceptable image reconstruction. PSNR is computed as:
\begin{equation}
PSNR = 10 \cdot \log_{10}\left(\frac{1}{MSE}\right)
\end{equation}

where $MSE$ is the mean squared error between the generated and reference normalized reflectance values, 
calculated across all six spectral bands.

\subsection{Quantitative Comparisons}

Table \ref{tab:inference_steps} summarizes the effect of the number of inference steps on the model performance. Notably, even with 3 inference steps the model achieves a decent baseline (SSIM = 0.738; SID = 0.039), illustrating the efficiency of linearized paths in Conditional Flow Matching. Performance steadily improves as the number of steps increases, with diminishing returns beyond 50 steps (SSIM: 0.912 at 50 steps vs. 0.908 at 100 steps). We thus select 50 steps to balance computational cost and accuracy.

\begin{table}[h]
\caption{Performance Metrics vs. Number of Inference Steps}
\label{tab:inference_steps}
\vskip 0.15in
\begin{center}
\begin{small}
\begin{sc}
\begin{tabular}{@{}lcccccc@{}}
\toprule
\textbf{Steps} & 1 & 3 & 5 & 10 & 50 & 100 \\
\midrule
SID & 0.285 & 0.039 & 0.0216 & 0.0194 & 0.018 & 0.012 \\
SSIM & 0.651 & 0.738 & 0.862 & 0.895 & 0.912 & 0.908 \\
PSNR & 23.3 & 28.5 & 29.7 & 29.9 & 31.8 & 30.5 \\
\bottomrule
\end{tabular}
\end{sc}
\end{small}
\end{center}
\vskip -0.1in
\end{table}

We evaluate our CFM approach against a conditional diffusion method \cite{DiffCR2024} and a traditional remote sensing fusion baseline (STARFM). For comparision, we chose number of inference steps as 50 for both CFM and diffusion models. Table \ref{tab:loss_comparison} shows that CFM outperforms alternatives in terms of SID, SSIM, and PSNR. These gains translate directly to higher-quality reconstructions in both downscaling (500m MODIS to 30m Landsat) and cloud gap-filling scenarios.

\begin{table}[H]
\caption{Comparison with Baseline Methods on Held-Out Scenes}
\label{tab:loss_comparison}
\vskip 0.15in
\begin{center}
\begin{small}
\begin{sc}
\begin{tabular}{lccc}
\toprule
\textbf{Method} & \textbf{SID} & \textbf{SSIM} & \textbf{PSNR} \\
\midrule
STARFM & 0.0481 & 0.852 & 28.6 \\
Diffusion & 0.0271 & 0.891 & 30.0 \\
\textbf{CFM} & \textbf{0.0186} & \textbf{0.912} & \textbf{31.8} \\
\bottomrule
\end{tabular}
\end{sc}
\end{small}
\end{center}
\vskip -0.1in
\end{table}

Lastly, we assess cloud imputation accuracy under different cloud coverage (10\%, 25\%, 50\%, and 75\%). Table \ref{tab:cloud_cover} demonstrates the efficacy of multi-sensor fusion: adding MODIS consistently yields lower SID and higher SSIM. This advantage becomes more pronounced as cloud coverage increases. For instance, at 75\% coverage, our method with MODIS exhibits lower SID and higher SSIM compared to the scenario without MODIS data, emphasizing the importance of leveraging coarse daily observations in heavily occluded conditions.

\begin{table}[H]
\caption{Performance vs. Cloud Cover (\%) With and Without MODIS Input}
\label{tab:cloud_cover}
\vskip 0.15in
\begin{center}
\begin{small}
\begin{sc}
\begin{tabular}{lcccc}
\toprule
\textbf{Cloud Cover (\%)} & \multicolumn{2}{c}{\textbf{With MODIS}} & \multicolumn{2}{c}{\textbf{Without MODIS}} \\
\midrule
 & SID & SSIM & SID & SSIM \\
\midrule
10 & 0.015 & 0.960 & 0.028 & 0.932 \\
25 & 0.032 & 0.921 & 0.056 & 0.884 \\
50 & 0.068 & 0.875 & 0.098 & 0.821 \\
75 & 0.071 & 0.812 & 0.167 & 0.723 \\
\bottomrule
\end{tabular}
\end{sc}
\end{small}
\end{center}
\vskip -0.1in
\end{table}

Together, these findings indicate that (1) our approach offers a robust framework for combining data from multiple sensors, (2) linearized flows enable faster, more efficient inference, and (3) incorporating MODIS observations in gap-filling further enhances resilience to occlusions  by providing additional temporal and spectral context.

\begin{figure}[ht]
    \vskip 0.2in
    \begin{center}
    \centerline{\includegraphics[width=\columnwidth]{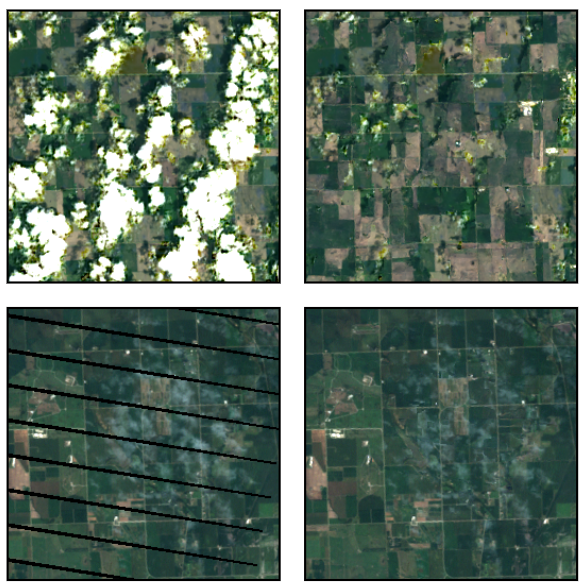}}
    \caption{Example of artifacts introduced by Quality Assessment misclassification. The images on the left show the original cloudy Landsat image, and the images on the right show resulting artifacts in the gap-filled output.}
    \label{fig:failure}
    \end{center}
    \vskip -0.2in
    \end{figure}

\section{Limitations}
While our proposed framework demonstrates strong performance in generating gap-free daily Landsat-like imagery, there remain several important limitations. First, our gap-filling strategy (Algorithm~\ref{alg:imputation}) relies on a mask that distinguishes clear pixels from contaminated ones. In our work, we utilize the quality assessment masks provided by Landsat level-2 processed products. In practice, these masks are prone to misclassification—particularly at cloud edges or shadows—which can introduce artifacts in the reconstructed outputs. As shown in Figure~\ref{fig:failure}, misclassifications can lead to visible artifacts and degraded image quality in the outputs. Advanced cloud and shadow detection algorithms could alleviate these artifacts.
Second, preprocessing of daily MODIS imagery involves temporal interpolation for missing or cloudy observations. However, linear or spline interpolation will perform poorly in the presence of cloud cover over extended time periods and extreme events (e.g., wildfires, floods, or snowfall), which may feature abrupt spectral changes. In such scenarios, the reconstruction will not reflect the real-world conditions.  Incorporating complementary modalities, such as Sentinel-1 SAR (Synthetic Aperture Radar) data and multiple remote sensing sources, may mitigate this shortcoming. However availability of newer earth observation datasets (e.g., Sentinel missions) is limited to the years after 2015.

\section{Conclusion and Future Work}

We presented a Conditional Flow Matching (CFM) model that fuses daily coarse-resolution MODIS imagery with Landsat observations to generate gap-free, high-resolution surface reflectance data.
We proposed integration of this model into a framework (SatFlow) to produce gap-free, Landsat-like imagery at regular intervals. This capability facilitates the generation of long-term remote sensing datasets, enhancing environmental monitoring and modeling applications. Our experimental results demonstrate that, particularly under high occlusion rates, the combined utilization of MODIS coarse data and Landsat composites allows reliable gap filling. In forthcoming work, we aim to extend the framework to include additional remote sensing sources such as Sentinel-2 optical imagery, VIIRS, and SAR data (Sentinel-1), aiming to further enhance robustness in cloudy or otherwise adverse conditions. We also plan to investigate efficient architectures derived from Vision Transformers (ViT) and Swin-UNet models, with the goal of achieving faster and better performing models capable of scaling to continental or global domains. Finally, we intend to quantify uncertainty in the generated reflectance maps, thereby providing reliability estimates for subsequent remote sensing analyses and decision-making.

\section*{Software and Data}

The software and dataset will be made available upon completion of review process.

\section*{Acknowledgements}

This work was supported by the United States Department of Agriculture (USDA) and the Oak Ridge Institute for Science and Education (ORISE) fellowship program. The findings and conclusions in this publication are those of the authors and should not be construed to represent any USDA determination or policy.

\section*{Impact Statement}

This paper presents work whose goal is to enhance earth observation with 
advanced techniques in Machine Learning. While our work may have various societal implications, 
we do not identify any that require specific discussion here.

\bibliography{example_paper}
\bibliographystyle{icml2025}




\end{document}